\documentclass[11pt]{article}
\usepackage{graphicx}
\usepackage{amsmath}
\usepackage{amssymb}
\usepackage{geometry}
\usepackage{url}
\usepackage{multirow}
\usepackage{subcaption}
\usepackage{authblk}
\usepackage[authoryear,round,sort]{natbib}
\title{\textbf{Spatio-Temporal Wildfire Spread Prediction in Canada using a Video Swin-Hybrid-U-Net and Satellite Imagery}}

\author[1,2*]{Maulik Srivastava}
\author[2]{Esha Saha}
\author[2,3]{Hao Wang}

\affil[1]{Faculty of Science, University of Alberta}

\affil[2]{Interdisciplinary Lab for Mathematical Ecology \& Epidemiology (ILMEE), University of Alberta}

\affil[3]{Department of Mathematical and Statistical Sciences, University of Alberta}

\affil[*]{Corresponding author. Email: maulik2@ualberta.ca}

\date{}

\begin{document}

\maketitle

\begin{abstract}
\textbf{Background:} Wildfires in Canada present increasing threats to ecosystems, communities, and infrastructure, demanding accurate forecasting tools to aid mitigation efforts. Existing models often lack scalability or fail to capture temporal dynamics effectively. \textbf{Aims:} This study aims to develop a deep learning framework tailored to Canadian wildfire spread prediction that captures spatio-temporal patterns in environmental data. \textbf{Methods:} We propose a U-Net architecture integrating a Video Swin Transformer encoder with a convolutional decoder to model three-day sequences of meteorological and environmental variables. Data are exclusively sourced from public repositories via Google Earth Engine, ensuring transparency and scalability. The model is trained and tested on a curated dataset of major Canadian wildfire events from 2014 to 2023. \textbf{Key results:} Our approach achieves strong predictive performance by effectively leveraging spatio-temporal attention to forecast next-day fire incidence maps. \textbf{Conclusions:} The model successfully captures complex wildfire dynamics unique to Canada’s landscape and temporal variability. \textbf{Implications:} This framework paves the way for advanced spatio-temporal wildfire forecasting research and operational applications using publicly accessible datasets. \end{abstract}

\section{Introduction}

The 2023 Canadian wildfire season was the most destructive on record, scorching approximately 15 million hectares of land \citep{ciffc2023}, an area larger than the country of England, and forcing the evacuation of approximately 232,000 people \citep{jones2024}. This unprecedented devastation underscores a critical reality: the wildfire regime is escalating across the boreal zone, as evidenced by extensive long-term satellite-based observations\citep{morenoruiz}. This increasing frequency and intensity, driven by a dynamic interplay of weather, topography, and fuel, present a formidable challenge to existing prediction and management efforts. For authorities to allocate firefighting resources effectively, issue timely evacuation orders, and protect critical infrastructure, they need accurate, spatially-explicit forecasts of a fire’s spread. 

The field of wildfire modeling has evolved from traditional physics-based simulators to deep learning methods that leverage remote sensing and environmental data. Classical models such as FARSITE \citep{finney1998} are grounded in empirical observations and physical principles, providing interpretable simulations, but often demand extensive parameter tuning and can be computationally intensive. In contrast, deep learning approaches have gained traction for their ability to learn directly from large-scale geospatial data \citep{jain2020}. 

Yet, effectively capturing the temporal dynamics of wildfire spread remains a critical challenge, particularly when adapting these models to the specific environmental conditions of regions like Canada. Recent work has shown that incorporating sub-daily weather observations into machine learning models can improve short-term fire forecasting, highlighting the importance of capturing short-term weather variability in wildfire prediction \citep{ardid2025}. Many current wildfire prediction models, however, are not optimized for the unique Canadian context \citep{wildfire-canada-risk}. Much of the leading research focuses on regions like the United States or the Mediterranean, where different climate patterns and ecosystems produce fire dynamics distinct from those in Canada’s vast boreal forests \citep{jain2020}. 

Furthermore, many advanced systems rely on proprietary data, creating barriers to access and hindering the reproducibility and scalability required for a Canada-wide operational tool \citep{wildfire_model_strategy}. This paper addresses this gap by training a model exclusively on Canadian wildfire events using publicly available data from Google Earth Engine (GEE), ensuring our methodology is transparent and reproducible. 

We introduce a novel deep learning framework for next-day wildfire spread prediction that makes two key contributions:
\begin{enumerate}
    \item \textbf{A Novel Architecture with Spatio-Temporal Attention:} Our work introduces a Video Swin-U-Net architecture that employs spatio-temporal attention, processing the input data as a unified 3D volume. This design allows the model to learn complex, long-range dependencies across both space and time simultaneously, effectively capturing the dynamic evolution of environmental conditions. This approach marks a fundamental departure from many existing models that rely on 2D convolutions or spatial-only attention mechanisms, which inherently limit their ability to truly model the progression of dynamic events like wildfires \citep{tran2015}. 
    \item \textbf{Deep Temporal Feature Learning:} Our model learns from a three-day sequence to understand how conditions evolve. Critically, key meteorological variables like wind and temperature are sampled multiple times per day. This fine-grained temporal resolution, which is rare in many existing wildfire datasets \citep{klock2023}, allows our model to capture transient diurnal patterns that are crucial drivers of fire behavior but are often missed by models using only daily data inputs. 
\end{enumerate}

By framing the problem as a video-to-image semantic segmentation task, our model provides a spatially explicit forecast of next-day fire presence, offering a powerful and intuitive tool for advancing research into spatio-temporal wildfire dynamics. This work represents a step forward in applying state-of-the-art computer vision techniques to the critical domain of environmental hazard prediction. 

\section{Materials and methods}

A robust and comprehensive dataset is fundamental to training a reliable wildfire prediction model. Our entire data pipeline is built upon publicly accessible satellite and reanalysis products available through Google Earth Engine (GEE). The dataset was curated to cover significant wildfire events across Canada from the summer of 2014 to 2023, encompassing a diverse set of fire behaviors across various ecosystems. 

\subsection{Data Acquisition and Preprocessing}
Our model ingests two categories of data: dynamic (time-varying) and static (time-invariant). The data is collected for contiguous Canada and processed to a uniform 500m spatial resolution and a 224x224 pixel dimension for each sample. Each data sample represents a localized fire event, identified by grouping fire points from satellite data that are within 5km of each other into a single active fire zone. These zones are enclosed in standardized square bounding boxes, ensuring each sample captures a distinct fire event for accurate modeling. 

\textbf{Dynamic Variables:} These variables are captured for each of the three days preceding the prediction target. Several are sampled at 6-hour intervals to provide high temporal resolution. 
\begin{itemize}
    \item \textbf{Fire Mask:} The historical fire locations are derived from the MODIS/MOD14A1 'FireMask' product, providing daily snapshots of fire presence \citep{nasa2021}. This is the primary feature indicating the fire's recent progression. 
    \item \textbf{Weather Data:} We source key weather parameters from the ECMWF/ERA5-LAND Hourly dataset \citep{munozsabater2021}, using bands for "2m air temperature" (temperature 2 meters above ground), "total precipitation" (accumulated rainfall), and "U/V components of 10m wind" (east-west and north-south wind speeds, used to derive wind speed and direction). Each band is aggregated into four daily measurements at 6-hour intervals to capture diurnal weather patterns. 
    \item \textbf{Soil Moisture:} Root zone soil moisture is obtained from the GLDAS/V022 dataset, providing a daily average value that is crucial for assessing landscape dryness \citep{beaudoing2020}. 
\end{itemize}

\textbf{Static Variables:} These variables are constant for any given location and thus remain consistent throughout each of the three days in a sample, providing essential context about the landscape. 
\begin{itemize}
    \item \textbf{Topography:} Elevation data is sourced from NRCan/CDEM \citep{nrcan2013}. From this, we derive Slope and Aspect using GEE's terrain algorithms. These factors heavily influence fire behavior \citep{pyne1996}. 
    \item \textbf{Vegetation:} The Normalized Difference Vegetation Index (NDVI), a proxy for fuel availability \citep{bajocco2015}, is derived from the NOAA/VIIRS/VNP13A1 product \citep{didandbarreto2018}. We use a mean composite of the 16 days prior to the event to represent the prevailing vegetation condition. 
    \item \textbf{Population Density:} Human activity, as indicated by population density, is correlated with wildfire spread \citep{syphard2007}. We use population density maps from the CIESIN/GPWv411 dataset, selecting the map corresponding to the year of the fire event (2010, 2015, or 2020) \citep{ciesin2018}. 
\end{itemize}

The core of our data pipeline is the transformation of the collected data into a structured spatio-temporal tensor suitable for our model. For each fire event, we create a tensor of shape $(T, C, H, W)$, where $T=3$ days, $C=23$ channels, and $H, W = 224$. The channel dimension $C$ for each day is a combination of 18 dynamic channels and 5 static channels:
\begin{itemize}
    \item \textbf{18 Dynamic Channels}: Fire Mask (1), Temperature (4), Precipitation (4), Wind Speed (4), Wind Direction (4), and Soil Moisture (1). Note that Wind Speed and Direction are calculated from the U/V components of 10m wind. 
    \item \textbf{5 Static Channels}: NDVI, Elevation, Slope, Aspect, and Population Density. 
\end{itemize}

The static channels are replicated for each of the three time steps, allowing the model to correlate dynamic changes with the underlying static landscape at every point in the sequence.\\

Before being fed to the model, the data undergoes channel-wise normalization. Topographic and vegetation features undergo min-max scaling, with elevation being log-scaled first to handle its wide range. Fire mask channels, being binary, are not normalized. All other dynamic channels are standardized using Z-score normalization calculated across the entire training dataset. This ensures that all features are on a comparable scale, which is crucial for stable training \citep{lecun2012efficient}. Finally, the dataset is split into training, testing, and validation sets. 

\subsection{Problem Statement}
\begin{figure}[h!]
    \centering
    \includegraphics[width=0.7\textwidth]{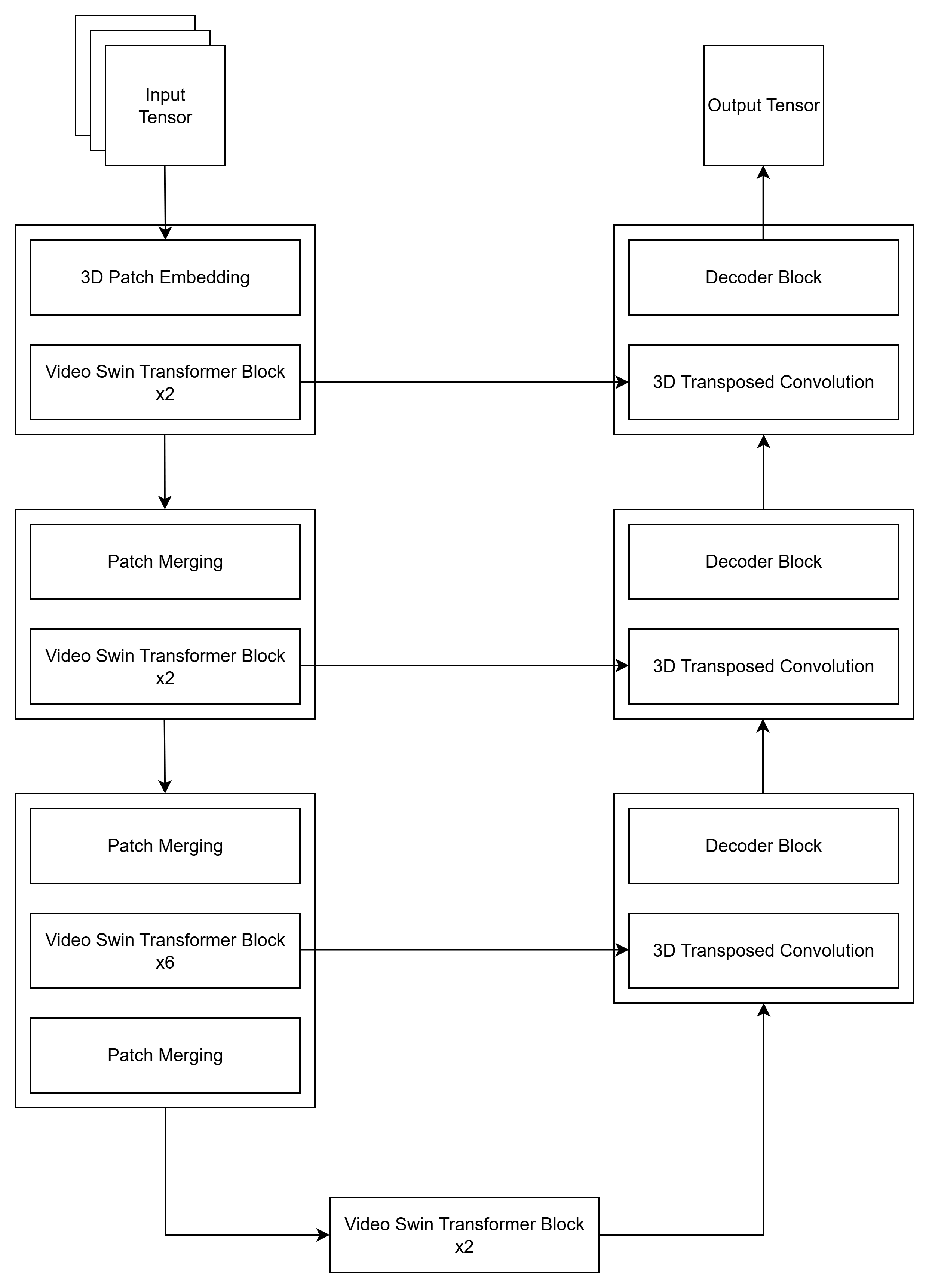} 
    \caption{\textbf{Overview of the Video Swin-U-Net Architecture.} The model takes a 3-day sequence of multi-channel data (left). The 3D Swin Transformer Encoder processes this 3D volume to extract hierarchical spatio-temporal features, producing feature maps at decreasing spatial resolutions (center-left). The Convolutional Decoder upsamples these features, integrating information from the encoder via skip connections (center-right). Temporal information is then aggregated by averaging the feature maps, and the final layers produce a 2D segmentation mask for the next day (right).}
    \label{fig:overview}
\end{figure}

The task of predicting wildfire spread is framed as a video-to-image semantic segmentation problem. The goal is to learn a mapping function, $f$, that takes a sequence of historical multi-channel environmental data and outputs a spatially explicit prediction of fire presence for the subsequent day. Formally, let the input be a tensor $X \in \mathbb{R}^{B \times T \times C \times H \times W}$ and the target output be a binary fire mask $Y \in \{0, 1\}^{B \times 1 \times H \times W}$ for the next day. The objective is to train the model $f$ such that the predicted mask $\hat{Y} = f(X)$ is as close as possible to the ground truth mask $Y$, as measured by a segmentation-focused loss function. Our proposed solution is a U-Net-like encoder-decoder architecture specifically designed to handle 3D spatio-temporal data (Figure \ref{fig:overview}). 
\begin{itemize}
    \item \textbf{Spatio-Temporal Encoder:} The encoder is a 3D Swin Transformer backbone \citep{liu2021video}. It ingests the spatio-temporal data sequence and generates a hierarchical feature representation. Its core strength lies in using shifted 3D window attention to efficiently model relationships across space and time. 
    \item \textbf{Convolutional Decoder:} The decoder progressively upsamples the feature representation from the encoder to the original spatial resolution using 3D transposed convolutions and skip connections to recover fine-grained spatial details \citep{ronneberger2015}. 
    \item \textbf{Loss Function:} The model is trained end-to-end using a weighted sum of Binary Cross-Entropy (BCE) and Dice Loss, which together mitigate the severe class imbalance inherent in wildfire segmentation tasks \citep{sudre2017}. 
\end{itemize}

\subsubsection{Spatio-Temporal Encoder}
To accurately predict fire spread, a model must understand not just the environment's state but also its evolution. Standard 2D models that flatten time into channels implicitly capture these changes but may not fully leverage the explicit temporal dynamics inherent in the data \citep{lahrichi2025predicting}. Our encoder addresses this by employing spatio-temporal attention to create rich representations directly from the full 3D data volume. 

\begin{figure}[h!]
    \centering
    \includegraphics[width=0.5\textwidth]{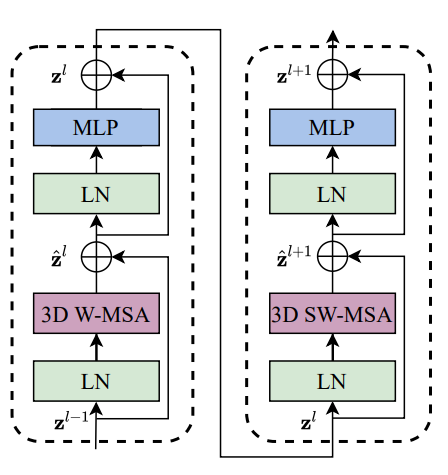}
    \caption{\textbf{Illustration of the Shifted Window Mechanism in a 3D Swin Transformer Block.} The block alternates between window-based multi-head self-attention (W-MSA) and shifted-window multi-head self-attention (SW-MSA) to enable cross-window connections while maintaining computational efficiency \citep{liu2021video}.}
    \label{fig:video_block}
\end{figure}

The encoder first divides the input into non-overlapping spatio-temporal patches and projects them into an embedding space. It then processes these embeddings through a series of 3D Swin Transformer blocks, which compute multi-head self-attention within local 3D windows (W-MSA) and use a shifted-window scheme (SW-MSA) to allow cross-window connections \citep{liu2021video} (Figure~\ref{fig:video_block}). The attention window size is dynamic, scaling with feature map resolution to provide a larger receptive field in early layers and finer attention in deeper ones. Between stages, a patch merging operation reduces spatial resolution by a factor of 2 and increases channel depth, producing a hierarchical feature pyramid \citep{lin2017feature} that captures information at multiple scales. Feature maps from each stage are stored for skip connections to the decoder. 

\subsubsection{Convolutional Decoder}
The decoder's role is to translate the encoder's abstract features back into a full-resolution segmentation map. Skip connections are vital for re-introducing high-resolution spatial details from the encoder, enabling precise localization in the final prediction \citep{ronneberger2015}. The decoder mirrors the encoder's structure, progressively upsampling features using 3D Transposed Convolutions. This method was chosen over alternatives like PixelShuffle, which in early tests led to checkerboard artifacts and overfitting \citep{aitken2017}. At each stage, the upsampled feature map is concatenated with the corresponding skip connection from the encoder and refined through 3D convolutional layers. Finally, the resulting 3D feature tensor is aggregated along the temporal dimension by averaging. This 2D map is then upscaled to the original resolution to produce the final single-channel logit map for segmentation. 

\subsubsection{Loss Function}
Wildfire segmentation faces severe class imbalance, with fire pixels being much rarer than non-fire pixels. A single loss function may struggle to handle both this imbalance and the spatial coherence of segmentation. We address this by combining two complementary loss functions. The total loss is a weighted sum of Weighted Binary Cross-Entropy (BCE) and Dice Loss, balanced by hyperparameters $\lambda_1$ and $\lambda_2$:
\[
L_{\text{Total}} = \lambda_1 \cdot L_{\text{BCE}} + \lambda_2 \cdot L_{\text{Dice}}.
\]
In our final model, we found that an equal weighting performed best, and thus set $\lambda_1 = 0.5$ and $\lambda_2 = 0.5$. Weighted BCE emphasizes accurate pixel-wise classification by assigning greater weight to the fire class. Dice Loss maximizes spatial overlap between prediction and ground truth, directly optimizing Intersection-over-Union (IoU):
\[
L_{\text{Dice}}(p, q) = 1 - \frac{2 \sum_{i=1}^{N} p_i q_i + \epsilon}{\sum_{i=1}^{N} p_i + \sum_{i=1}^{N} q_i + \epsilon},
\]
where \(p\) is the predicted probability, \(q\) the ground truth mask, \(N\) the pixel count, and \(\epsilon\) avoids division by zero. This combined loss enables balanced classification of rare fire pixels while encouraging spatially coherent segmentation \citep{sudre2017}. 

\section{Results}
\subsection{Experimental Setup}

\noindent\textbf{Dataset:} We use our custom-curated dataset of Canadian wildfires from 2014 to 2023, as described in the Results section. The dataset is first split into a fixed, held-out 20\% test set. The remaining 80\% is used for training. To ensure robust results and account for statistical variance, all experiments were repeated 5 times using 5 different random seeds. For each run, the 80\% training data was randomly re-split into new training and validation sets based on that run's seed. The final, best-performing model from each of the 5 runs was then evaluated on the single, fixed test set. All metrics reported in Table \ref{tab:results} are the \textbf{mean $\pm$ standard deviation} of these 5 test set scores. 
\medskip

\noindent\textbf{Baselines:} To evaluate the effectiveness of our proposed architecture, we compare it against several strong baseline models, including a standard 2D U-Net \citep{ronneberger2015}, a 2D Swin U-Net \citep{cao2022} which uses a Swin Transformer encoder, and a pure Video Swin U-Net which employs 3D Swin Transformer blocks in both its encoder and decoder, to isolate the effectiveness of our hybrid design. In both 2D baselines, temporal data was concatenated along the channel axis. 
\medskip

\noindent\textbf{Evaluation Metrics:} The choice of metrics is critical for this task due to the \textbf{severe class imbalance} where non-fire pixels vastly outnumber fire pixels. 
\begin{itemize}
    \item \textbf{F1-Score, Precision, \& Recall:} These metrics are essential for imbalanced classification. Precision measures the accuracy of positive predictions, minimizing false alarms. Recall measures the model's ability to identify all true fire pixels, crucial for detecting active fire spread. The F1-Score, the harmonic mean of precision and recall, provides a balanced measure of performance when both false positives and false negatives are costly \citep{powers2020}. 
\end{itemize}

\subsection{Training and Implementation Details}

The model is trained end-to-end using the AdamW optimizer with a learning rate of 0.0001 \citep{loshchilov2017}. A Cosine Annealing Warm Restarts schedule is employed to adjust the learning rate during training, which helps in escaping local minima and finding a more robust solution \citep{loshchilov2016}. To accelerate training and reduce GPU memory consumption, we utilize automatic mixed-precision training with a gradient scaler. Gradient clipping with a max norm of 1.0 is also applied to prevent exploding gradients and stabilize the training process \citep{pascanu2013}. All models were trained for 150 epochs with a batch size of 8 on an NVIDIA RTX 3070 GPU using the PyTorch framework. Our proposed model uses an embedding dimension of 96, and the 3D Swin Transformer has stage depths of [2, 2, 6, 2] with a corresponding number of attention heads of [3, 6, 12, 24]. The patch size is (1, 4, 4) and the initial attention window size is (3, 8, 8), which dynamically changes, as stated in the Spatio-Temporal Encoder section. 

\subsection{Scores}

The performance of our proposed Hybrid Video Swin U-Net model was comprehensively evaluated on the test set, combining quantitative, qualitative, and topographical analyses to provide a holistic understanding of its predictive capabilities. As shown in Table \ref{tab:results}, our model demonstrates superior performance across all key segmentation metrics when compared against established baselines. The Hybrid Video Swin U-Net achieved an F1-Score of \textbf{0.6550}, a precision of \textbf{0.6328}, and a recall of \textbf{0.6791}, outperforming the next-best model (Video Swin U-Net) by a significant margin. 

\begin{table}[h!]
    \centering
    \caption{Performance comparison on the test set. Our proposed model is compared against three baselines. All metrics are reported as the mean $\pm$ standard deviation across 5 independent runs with different random seeds.}
    \label{tab:results}
    \begin{tabular}{l|ccc}
        \hline
        \textbf{Model} & \textbf{F1-Score} & \textbf{Precision} & \textbf{Recall} \\
        \hline
        2D U-Net & $0.3647 \pm 0.015$ & $0.3382 \pm 0.018$ & $0.3957 \pm 0.012$ \\
        2D Swin U-Net & $0.4528 \pm 0.012$ & $0.4328 \pm 0.014$ & $0.4749 \pm 0.011$ \\
        Video Swin U-Net & $0.6003 \pm 0.008$ & $0.6196 \pm 0.007$ & $0.5823 \pm 0.009$ \\
        \hline

        \textbf{Hybrid Video Swin U-Net} & \textbf{$0.6550 \pm 0.006$} & \textbf{$0.6328 \pm 0.007$} & \textbf{$0.6791 \pm 0.005$} \\
        \hline
    \end{tabular}
\end{table}

\begin{figure}[h!]
    \centering
    \begin{tabular}{cc}
        \includegraphics[width=0.45\textwidth]{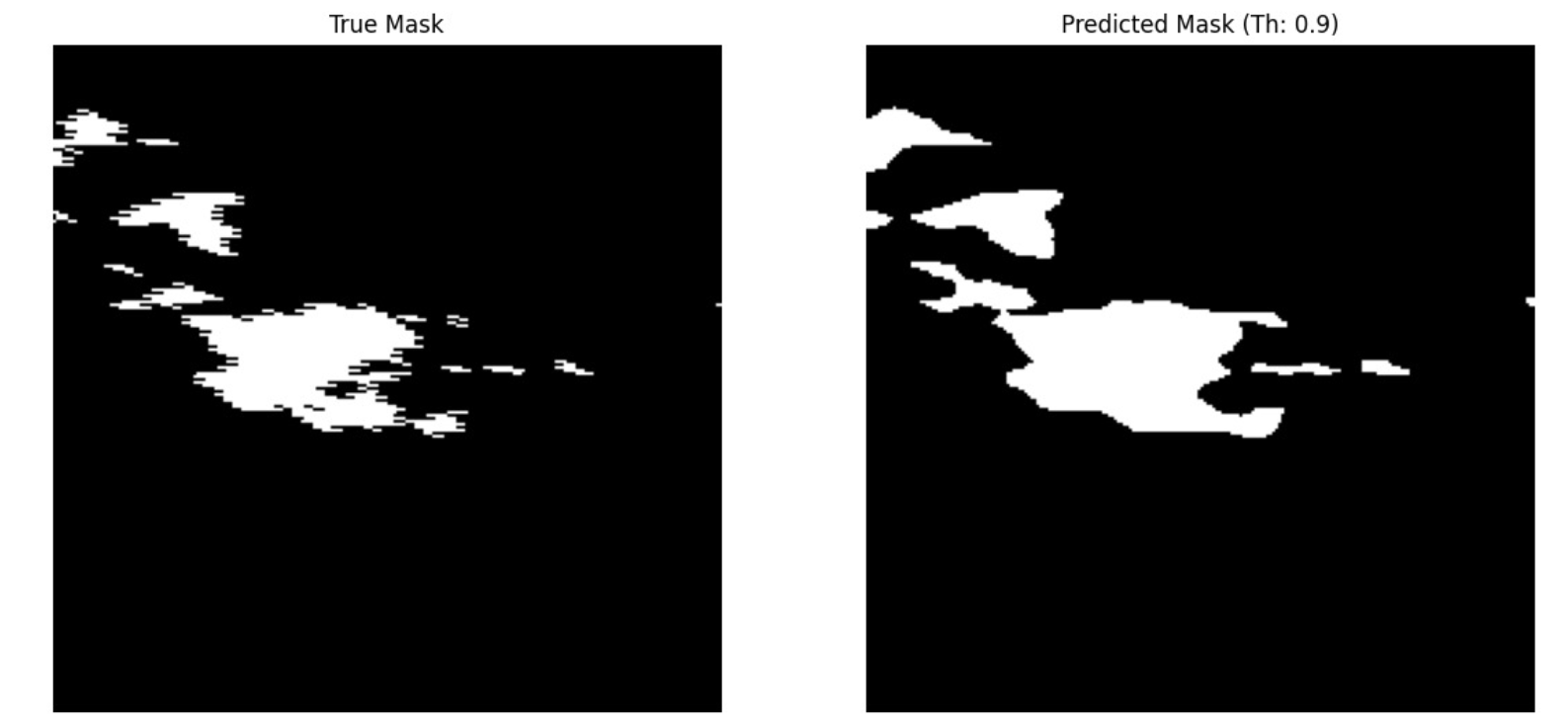} &
        \includegraphics[width=0.45\textwidth]{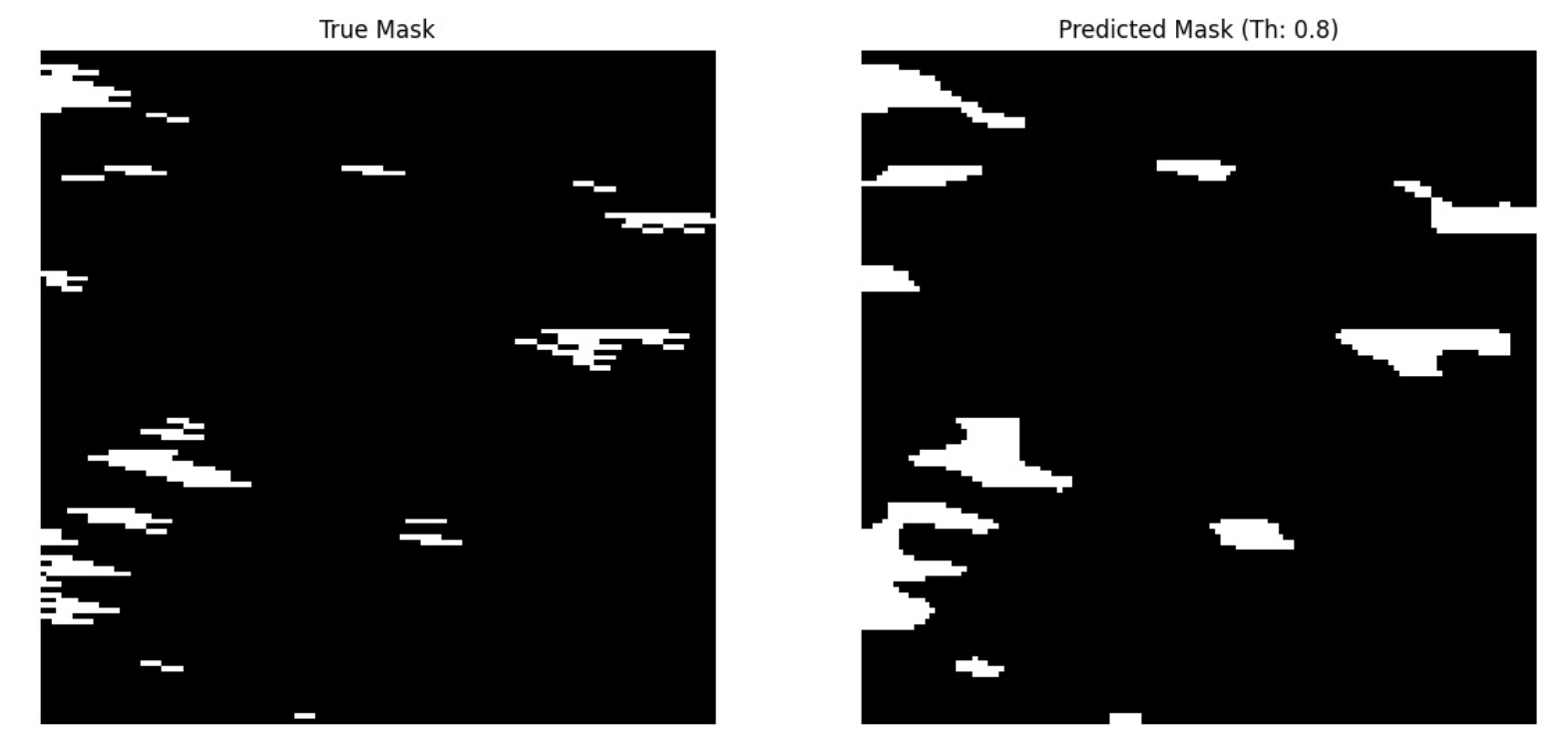} \\
        \includegraphics[width=0.45\textwidth]{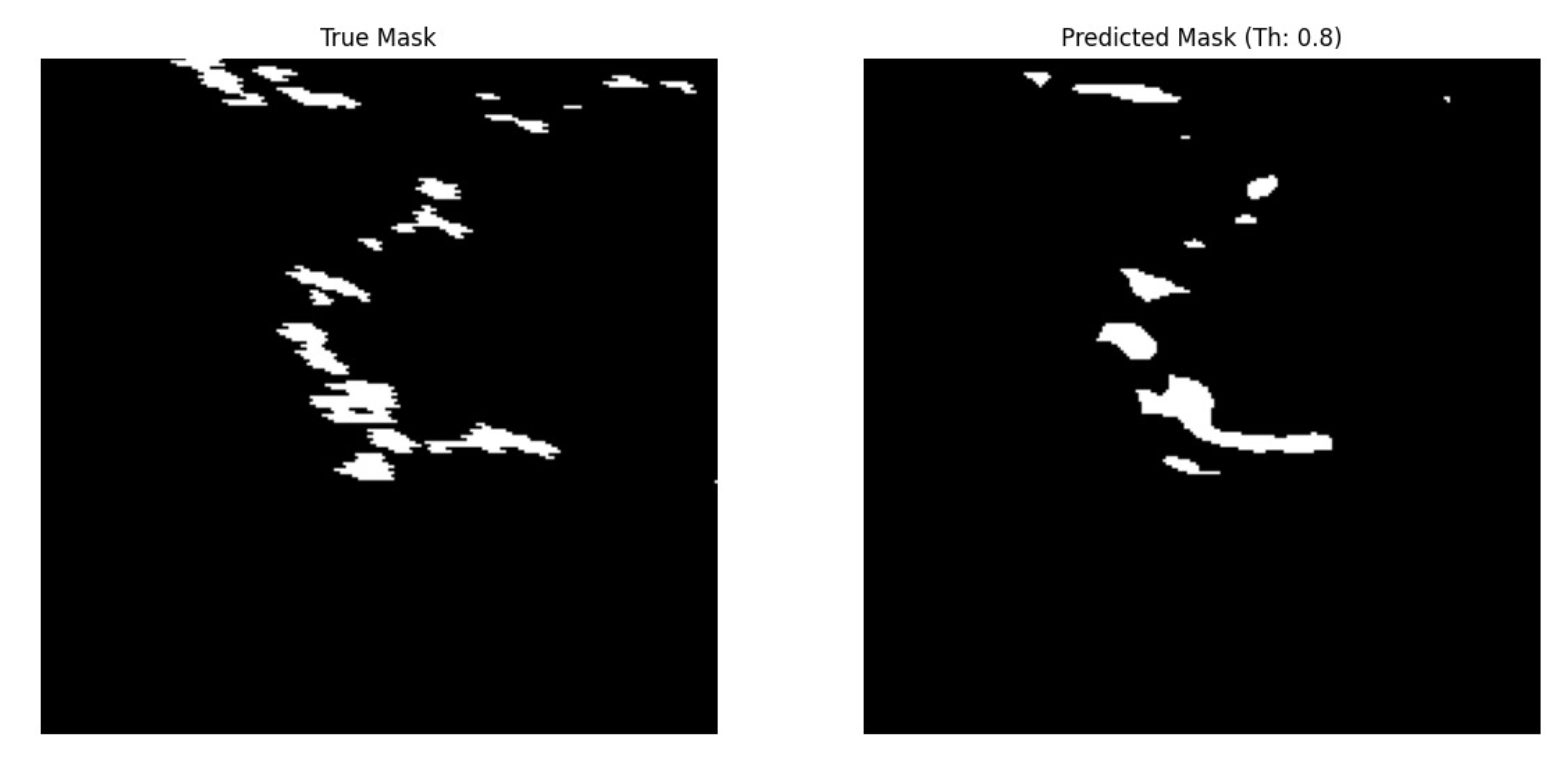} &
        \includegraphics[width=0.45\textwidth]{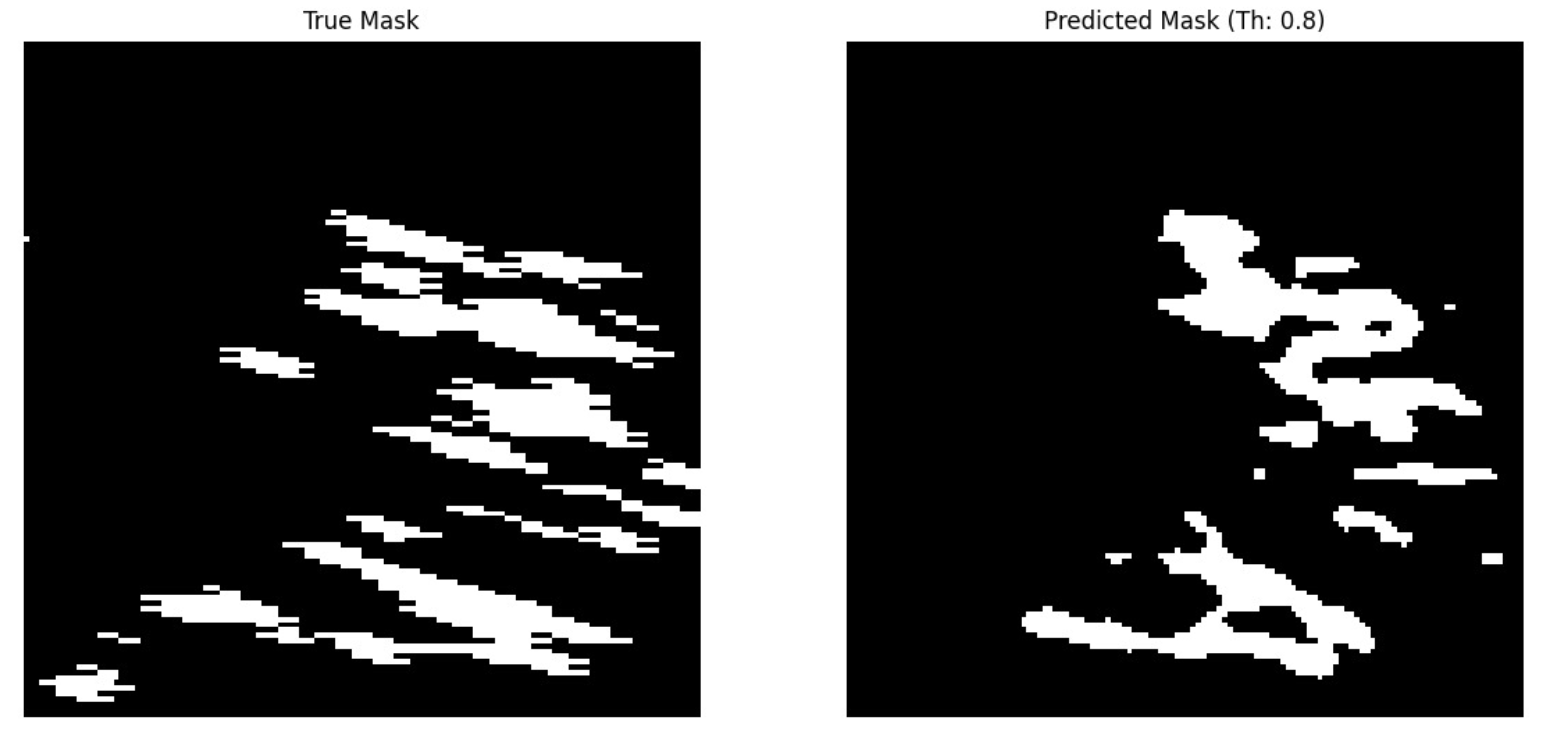} \\
    \end{tabular}
    \caption{A visual comparison of predicted fire spread and ground truth for samples from the test set.}
    \label{fig:qualitative_results}
\end{figure}

To complement the aggregate metrics, we conducted a qualitative visual inspection and a detailed topographical error analysis. Figure \ref{fig:qualitative_results} presents a representative example from the test set, comparing our model's predictions to the ground truth. The visualization highlights the model's ability to capture the complex shapes and boundaries of fire spread. 

To understand how performance varies with key environmental factors, we conducted analyses based on both static topographical features and dynamic soil moisture values. Pixels were sorted independently by normalized Digital Elevation Model (DEM) and Slope, as well as by soil moisture values for each of the three input days. We then created five bins, with the boundaries of each bin set so that each contains an equal number of pixels. The F1-Score was then calculated within each bin for the topographical features (Table \ref{tab:f1_topo_bins}) and F1-Score, precision, and recall computed for soil moisture bins (Table \ref{tab:f1_sm_bins}). 

\begin{table}[h!]
    \centering
    \caption{F1-Score by Digital Elevation Model (DEM) and Slope Bins.}
    \label{tab:f1_topo_bins}
    \begin{tabular}{l|c|l|c|l}
        \hline
        \textbf{Percentile Bin} & \textbf{DEM Range} & \textbf{DEM F1-Score} & \textbf{Slope Range} & \textbf{Slope F1-Score} \\
        \hline
        0-20th & (0.0000, 0.0758] & 0.6573 & (0.0000, 0.0028] & 0.6189 \\
        20-40th & (0.0758, 0.0966] & 0.6201 & (0.0028, 0.0075] & 0.6299 \\
        40-60th & (0.0966, 0.1200] & 0.6299 & (0.0075, 0.0156] & 0.6351 \\
        60-80th & (0.1200, 0.1630] & 0.6408 & (0.0156, 0.0329] & 0.6688 \\
        80-100th & (0.1630, 1.0000] & 0.7269 & (0.0329, 1.0000] & 0.7223 \\
        \hline
    \end{tabular}
\end{table}

The analysis, as detailed in Table \ref{tab:f1_topo_bins}, revealed distinct trends. For slope, we observed a strong, positive correlation between terrain steepness and model performance. The F1-Score consistently improved from \textbf{0.6189} on the flattest terrain to \textbf{0.7223} on the steepest terrain. In contrast, the model's performance relative to elevation showed a more complex, non-linear relationship. The F1-Score was highest in the lowest and highest elevation bins (\textbf{0.6573} and \textbf{0.7269}, respectively) but showed a notable dip in performance for mid-range elevations (F1-Score of \textbf{0.6299}). 

In contrast to the static topographical features, the analysis of the dynamic soil moisture variable (Table \ref{tab:f1_sm_bins}) indicates that the model's performance is notably stable across all moisture conditions. For all three input days, the F1-Score remains consistently high, with all bins scoring between 0.64 and 0.68. This robustness suggests the model has successfully learned to predict fire spread regardless of whether the ground is extremely dry (e.g., Day 1 bin [-4.22, -0.58)) or moderately wet (e.g., Day 3 bin [0.88, 2.83]). 

\begin{table}[h!]
    \centering
    \caption{F1-Score across Soil Moisture Bins for each Input Day.}
    \label{tab:f1_sm_bins}
    \begin{tabular}{l|c|c}
        \hline
        \textbf{Day} & \textbf{Bin Range} & \textbf{F1-Score} \\
        \hline
        \multirow{5}{*}{Day 1} & [-4.22, -0.58) & 0.6712 \\
        & [-0.58, -0.16) & 0.6549 \\
        & [-0.16, 0.22) & 0.6510 \\
        & [0.22, 0.87) & 0.6797 \\
        & [0.87, 2.88) & 0.6471 \\
        \hline
        \multirow{5}{*}{Day 2} & [-4.24, -0.58) & 0.6695 \\
        & [-0.58, -0.16) & 0.6528 \\
        & [-0.16, 0.22) & 0.6599 \\
        & [0.22, 0.87) & 0.6782 \\
        & [0.87, 2.87) & 0.6439 \\
        \hline
        \multirow{5}{*}{Day 3} & [-4.26, -0.58) & 0.6632 \\
        & [-0.58, -0.14) & 0.6578 \\
        & [-0.14, 0.24) & 0.6614 \\
        & [0.24, 0.88) & 0.6776 \\
        & [0.88, 2.83) & 0.6457 \\
        \hline
    \end{tabular}
\end{table}

\section{Discussion}

Our primary finding is that the proposed Hybrid Video Swin-U-Net architecture achieves significantly higher F1-Score, precision, and recall compared to baseline models on our curated dataset of Canadian wildfires. The Hybrid Video Swin U-Net achieved an F1-Score of 0.6550, a precision of 0.6328, and a recall of 0.6791, outperforming the next-best model, the Video Swin U-Net, by a significant margin. Specifically, our model's F1-Score of 0.6550 exceeded that of the Video Swin U-Net (0.6003) by approximately 5 percentage points and the best spatial attention model, the 2D Swin U-Net (0.4528), by over 20 percentage points. 

The performance of our model across all key segmentation metrics demonstrates that our hybrid approach effectively leverages spatio-temporal attention to model the evolving environmental context critical to wildfire spread prediction, resulting in a more balanced and accurate identification of fire-affected areas. These quantitative results are further supported by a qualitative visual inspection, as shown in Figure \ref{fig:qualitative_results}, which highlights our model's ability to capture the intricate shapes and boundaries of fire spread, producing predictions with smoother contours compared to the jagged boundaries often found in the ground truth masks. This characteristic may be attributed to the model's convolutional upsampling process, suggesting it learns a more generalized representation of the fire front. To further scrutinize the model's spatial accuracy, the Appendix provides extended visualizations explicitly mapping true positives, false positives, and false negatives across additional test samples. 

Our architectural design differs from existing approaches in several ways. For instance, the Swin-Unet-based model by \citet{lahrichi2025predicting} handles multi-day inputs by concatenating temporal slices along the channel dimension. While this method incorporates historical data, the temporal structure and evolution of environmental conditions are modeled implicitly through channel concatenation rather than explicitly through dedicated spatio-temporal mechanisms. Other models, such as ASUFM \citep{li2023} and MA-Net \citep{shadrin2024}, focused on spatial attention mechanisms or were trained using only a single day of input data, which inherently prevents them from learning from temporal sequences. Our approach, which processes the input as a unified 3D video volume, is fundamentally designed to overcome this limitation by using spatio-temporal attention to learn dependencies across both space and time simultaneously. 

More advanced temporal models have also been proposed. \citet{marjani2023} developed the FirePred model, which employs a recurrent neural network (ConvLSTM) to process a two-day sequence of historical data, demonstrating the value of sequential temporal modeling. More recently, \citet{zhao2025near} adapted the AR-SwinUNETR architecture from 3D medical imaging by treating the temporal axis as a third spatial dimension. Our methodology builds upon these temporal modeling insights while introducing a key distinction: By representing the input as sequential video frames, our Video Swin-U-Net architecture applies specialized spatio-temporal attention mechanisms that learn temporal dependencies directly and independently of the spatial representation. This design is better suited to capture the sequential, chronological progression of wildfire dynamics across multiple days and at high intra-day resolutions. 

The robustness of our results is also rooted in our dataset's unique characteristics. While the WildfireSpreadTS dataset used by \citet{lahrichi2025predicting} focuses on the USA and the FirePred model \citep{marjani2023} was developed for British Columbia, our dataset covers the full extent of contiguous Canada and is constructed entirely using public data from Google Earth Engine (GEE). Furthermore, our dataset's high temporal granularity---sampling temperature, precipitation, and wind variables four times per day---provides a richer signal for capturing the transient diurnal patterns that are essential for modeling wildfire dynamics, an advantage over models that rely on inputs sampled at lower daily resolutions. 

Beyond the aggregate performance metrics, the practical utility of a wildfire prediction model in real-world scenarios hinges on its robustness across diverse environmental contexts. To this end, we conducted a granular analysis to investigate how our model's accuracy varies with critical dynamic (soil moisture) and static (topography) environmental factors. This provides crucial insights into the model's generalizability, its strengths under specific conditions, and its capacity to capture the complex influence of these fundamental drivers on fire spread, thereby informing future model refinements and applications. 

The soil moisture-related performance analysis provides important insights into the model’s handling of fuel availability, which is a fundamental driver of wildfire behavior. Soil moisture is recognized as a key proxy for fuel dryness and combustibility, with drier soils indicating more flammable fuels. Our findings show that model F1-Scores, precision, and recall values remain relatively stable across the range of soil moisture bins for Day 1 through Day 3 inputs. Slightly higher performance is observed in intermediate soil moisture bins (around normalized values of 0.22 to 0.87), suggesting the model effectively discerns fire risk in conditions where fuels are neither extremely dry nor overly wet. The drop in performance at the highest moisture bins aligns with physical expectations, as wetter fuels reduce fire likelihood, inherently lowering the number of fire pixels and increasing class imbalance challenges for the model. At the lowest moisture levels, the model maintains competitive prediction quality, consistent with an ability to detect fire presence under highly flammable conditions. These results underscore the value of integrating temporally resolved soil moisture data to capture changing fuel availability and its impact on wildfire progression. 

Topographical analysis further elucidates how static landscape features affect prediction accuracy. The model’s F1-Score consistently increases with slope steepness, improving from approximately 0.6189 in the flattest terrain bin to 0.7223 in the steepest. This suggests that fire spread behavior on slopes, especially rapid upslope movement, may be more spatially coherent and thus easier for the model to predict, highlighting the utility of including slope as a contextual feature. Conversely, elevation shows a non-linear influence on performance, with highest accuracy at low and high elevation extremes, and a dip in mid-range elevations. This relationship likely reflects complex interactions with vegetation types, microclimates, and wind patterns varying by altitude, all affecting fire dynamics and prediction challenge. 

A key limitation of our approach is its dependence on the quality and availability of public satellite data. Optical sensors like MODIS are susceptible to cloud cover, which can lead to missing fire-mask data. This may impact predictive accuracy for certain events where data is obscured, a common challenge for remote sensing-based models. Another limitation lies in our data representation strategy where static channels are replicated across all input time steps. While ensuring consistent contextual information, this redundancy may introduce inefficiencies in feature learning and potentially limit the model’s ability to extract more compressed representations of unchanging landscape features . 

Despite these constraints, the principal strength of this work is its purpose-built architecture for learning from dynamic data. The entire framework’s reliance on a publicly-sourced data pipeline further ensures that the approach is transparent, reproducible, and easy to adopt. The interpretation of these results is clear: for dynamic environmental forecasting, how a model processes time is paramount. Models that rely primarily on 2D convolutions or spatial-only attention mechanisms face inherent challenges in distinguishing temporal sequencing from spatial features. Our work demonstrates that explicit spatio-temporal attention offers a robust mechanism for handling the complexities of time-series geospatial data. 

\section{Conclusion}

In this paper, we introduced a novel deep learning framework, the Video Swin-U-Net, for predicting wildfire spread in Canada. By interpreting the task as a video-to-image segmentation problem, our model effectively leverages spatio-temporal attention  and data from public satellites to generate spatially explicit forecasts. The core of our model is a 3D Swin Transformer encoder that excels at capturing long-range dynamic features, combined with an efficient convolutional decoder. Our approach demonstrates the significant potential of modern computer vision architectures to address critical challenges in environmental science. 

Future work could proceed in several exciting directions. Exploring longer prediction horizons (e.g., 3 or 5 days) for fire spread could allow the model to learn about longer-term climate signals. Incorporating detailed land cover type maps alongside NDVI could further enhance the representation of fuel availability, providing a more comprehensive understanding of combustible materials. Investigating more sophisticated methods for integrating static data is another promising direction. This could involve exploring alternative architectures, such as a separate processing pathway for static features or more advanced fusion techniques, to efficiently incorporate time-invariant landscape context without redundancy. Another promising direction is exploring different attention mechanisms or decoder designs. Finally, developing methods to quantify the model's predictive uncertainty would be a valuable step towards building operational trust in such forecasting systems. 

\clearpage
\appendix

\section{Appendix. Additional Performance Visualizations}
\label{sec:appendix_visuals}

\begin{figure}[h!]
    \centering
    
    \begin{subfigure}{0.32\textwidth}
        \includegraphics[width=\textwidth]{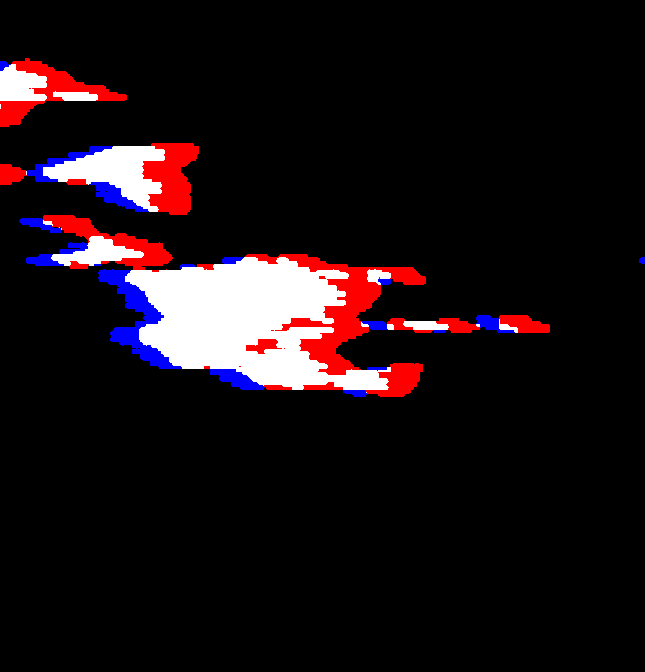}
        \caption{Sample 1}
    \end{subfigure}
    \hfill
    \begin{subfigure}{0.32\textwidth}
        \includegraphics[width=\textwidth]{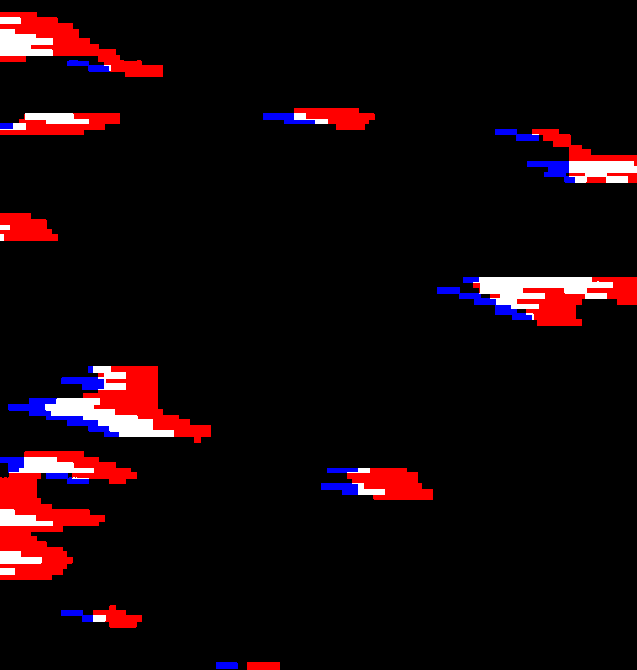}
        \caption{Sample 2}
    \end{subfigure}
    \hfill 
    \begin{subfigure}{0.32\textwidth}
        \includegraphics[width=\textwidth]{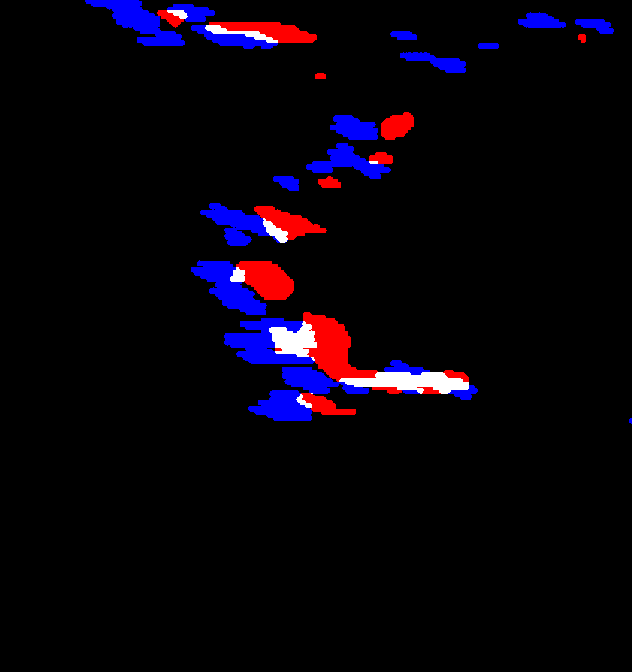}
        \caption{Sample 3}
    \end{subfigure}

    \vspace{5mm} 
    
    \begin{subfigure}{0.32\textwidth}
        \includegraphics[width=\textwidth]{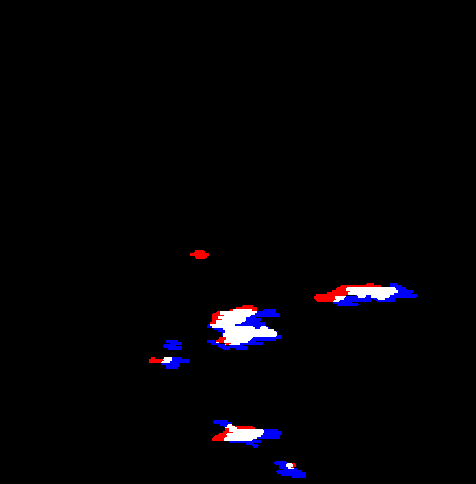}
        \caption{Sample 4}
    \end{subfigure}
    \hfill
    \begin{subfigure}{0.32\textwidth}
        \includegraphics[width=\textwidth]{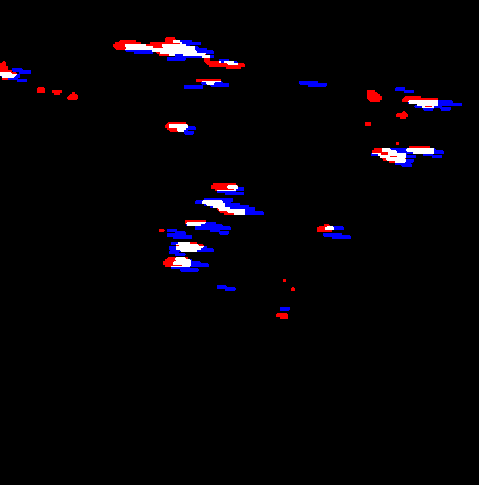}
        \caption{Sample 5}
    \end{subfigure}
    \hfill
    \begin{subfigure}{0.32\textwidth}
        \includegraphics[width=\textwidth]{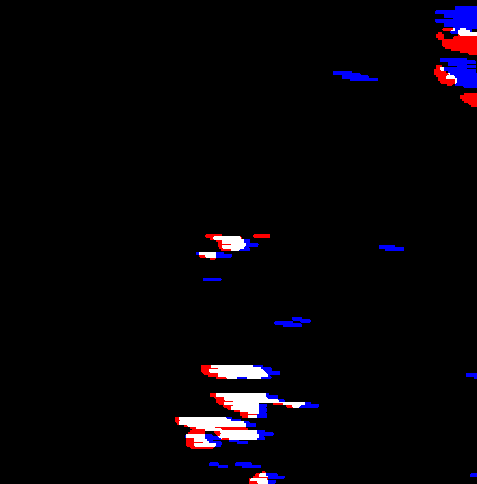}
        \caption{Sample 6}
    \end{subfigure}

    \vspace{5mm}
  
    \begin{subfigure}{0.32\textwidth}
        \includegraphics[width=\textwidth]{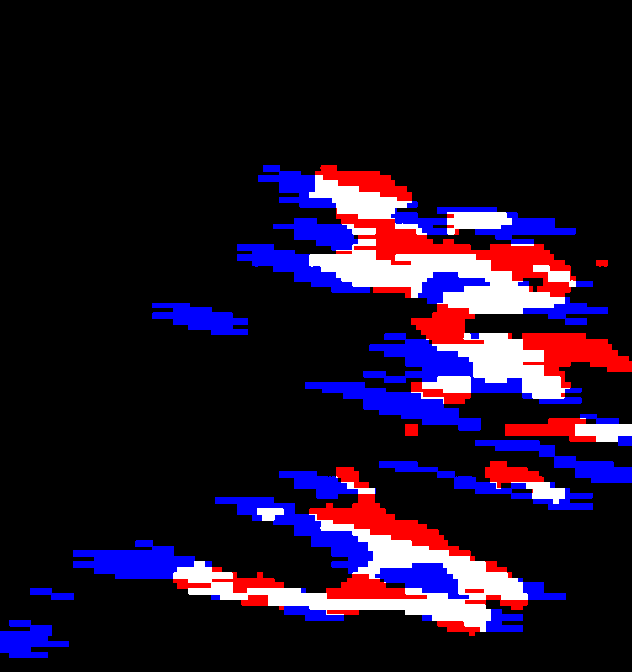}
        \caption{Sample 7}
    \end{subfigure}
    
    \caption{
        Additional visualizations for seven test samples, illustrating model performance. Pixels are color-coded to show agreement and disagreement with the ground truth:
        \textbf{White} = True Positive (Correctly predicted fire). \textbf{Red} = False Positive (Over-prediction).
        \textbf{Blue} = False Negative (Missed fire).
    }
    \label{fig:qualitative_grid}
\end{figure}

\section*{Figure Legends}
\setlength{\parindent}{0pt}
\textbf{Figure \ref{fig:overview}: Overview of the Video Swin-U-Net Architecture.} The model takes a 3-day sequence of multi-channel data (left). The 3D Swin Transformer Encoder processes this 3D volume to extract hierarchical spatio-temporal features, producing feature maps at decreasing spatial resolutions (center-left). The Convolutional Decoder upsamples these features, integrating information from the encoder via skip connections (center-right). Temporal information is then aggregated by averaging the feature maps, and the final layers produce a 2D segmentation mask for the next day (right). 

\textbf{Figure \ref{fig:video_block}: Illustration of the Shifted Window Mechanism in a 3D Swin Transformer Block.} The block alternates between window-based multi-head self-attention (W-MSA) and shifted-window multi-head self-attention (SW-MSA) to enable cross-window connections while maintaining computational efficiency \citep{liu2021video}. 

\textbf{Figure \ref{fig:qualitative_results}:} A visual comparison of predicted fire spread and ground truth for samples from the test set.

\section*{Acknowledgments}
We thank Amit Chakraborty (Interdisciplinary Lab for Mathematical Ecology \& Epidemiology, Department of Mathematical and Statistical Sciences, University of Alberta) for his support and feedback during this project.

\section*{Data Availability Statement}
All data used in this study are publicly available. The datasets were sourced from Google Earth Engine (GEE), including the MODIS/MOD14A1 fire product \citep{nasa2021}, ECMWF/ERA5-LAND hourly data \citep{munozsabater2021}, GLDAS/V022 soil moisture data \citep{beaudoing2020}, NRCan/CDEM topography data \citep{nrcan2013}, NOAA/VIIRS/VNP13A1 vegetation data \citep{didandbarreto2018}, and CIESIN/GPWv411 population data \citep{ciesin2018}. The full list of datasets is detailed in the Data Acquisition and Preprocessing section.

\section*{Conflicts of Interest}
The authors declare no conflicts of interest.

\section*{Declaration of Funding}
This work was supported by the Science Experiential Skills Advantage Award (Maulik Srivastava); the Grant Notley Postdoctoral Fellowship (Esha Saha); and the Senior Canada Research Chair program (Hao Wang).

\end{document}